\documentclass[letterpaper, 10 pt, conference]{ieeeconf}

\IEEEoverridecommandlockouts

\usepackage{etextools}
\usepackage{amssymb}
\usepackage{float}
\usepackage{makeidx}
\usepackage{amsmath}
\usepackage{bbm}
\usepackage{epsfig}
\usepackage{epsf}
\usepackage{psfrag}
\usepackage{verbatim}
\usepackage{color}
\usepackage{multirow}
\usepackage{tabularx}
\usepackage{booktabs}
\usepackage[tight,footnotesize]{subfigure}
\usepackage{array}
\usepackage{soul}
\usepackage{footnote}
\usepackage{cite}
\usepackage{dblfloatfix}

\usepackage{color, colortbl}
\usepackage[colorlinks,bookmarksnumbered,citecolor=orange,urlcolor=orange]{hyperref}
\usepackage{graphicx}
\graphicspath{{Figures}}
\DeclareGraphicsExtensions{.pdf,.png}
\usepackage{bigstrut}
\usepackage[english]{babel}

\usepackage{csquotes}

\usepackage[T1]{fontenc}
\usepackage{algorithm, setspace}
\usepackage{algpseudocode}
\usepackage{url}
\usepackage{multirow}
\usepackage{float}
\usepackage{xcolor}
\usepackage{hyperref}
 \hypersetup{
     colorlinks=true,
     linkcolor=orange,
     filecolor=orange,
     citecolor=orange,      
     urlcolor=orange,
     }

\newcommand{\p}[1]{\smallskip \noindent \textbf{{#1}.}}
\newcommand{\eq}[1]{Equation~(\ref{eq:#1})}
\newcommand{\fig}[1]{Figure~\ref{fig:#1}}
\usepackage{balance}
\usepackage{bbold}

\title{\LARGE
Towards Robots that Influence Humans over Long-Term Interaction
}

\author{Shahabedin Sagheb$^1$, Ye-Ji Mun$^2$, Neema Ahmadian$^1$, Benjamin A. Christie$^1$ \\ Andrea Bajcsy$^3$, Katherine Driggs-Campbell$^2$, and Dylan P. Losey$^1$
\thanks{$^1$Collaborative Robotics Lab (\href{https://collab.me.vt.edu/}{Collab}), Dept. of Mechanical Engineering, Virginia Tech, Blacksburg, VA 24061.}
\thanks{$^{2}$Human-Centered Autonomy Lab (\href{https://thehcalab.web.illinois.edu/}{HCAL}), Dept. of Electrical and Computer Engineering, University of Illinois at Urbana-Champaign, IL 61801.}
\thanks{$^{3}$Dept. of Electrical Engineering and Computer Science, UC Berkeley, CA 94709. Corresponding author's email: \texttt{shahab@vt.edu}}
}

\begin{document}
\maketitle

\begin{abstract}

When humans interact with robots influence is inevitable. Consider an autonomous car driving near a human: the speed and steering of the autonomous car will affect how the human drives. Prior works have developed frameworks that enable robots to influence humans towards desired behaviors. But while these approaches are effective in the short-term (i.e., the first few human-robot interactions), here we explore \textit{long-term} influence (i.e., repeated interactions between the same human and robot). Our central insight is that humans are dynamic: people adapt to robots, and behaviors which are influential now may fall short once the human learns to anticipate the robot's actions. With this insight, we experimentally demonstrate that a prevalent game-theoretic formalism for generating influential robot behaviors becomes less effective over repeated interactions. Next, we propose three modifications to Stackelberg games that make the robot's policy both influential and unpredictable. We finally test these modifications across simulations and user studies: our results suggest that robots which purposely make their actions harder to anticipate are better able to \textit{maintain} influence over long-term interaction. See videos here: \url {https://youtu.be/ydO83cgjZ2Q}

\end{abstract}


\section{Introduction}

\begin{figure}[t]
	\begin{center}
		\includegraphics[width=1\columnwidth]{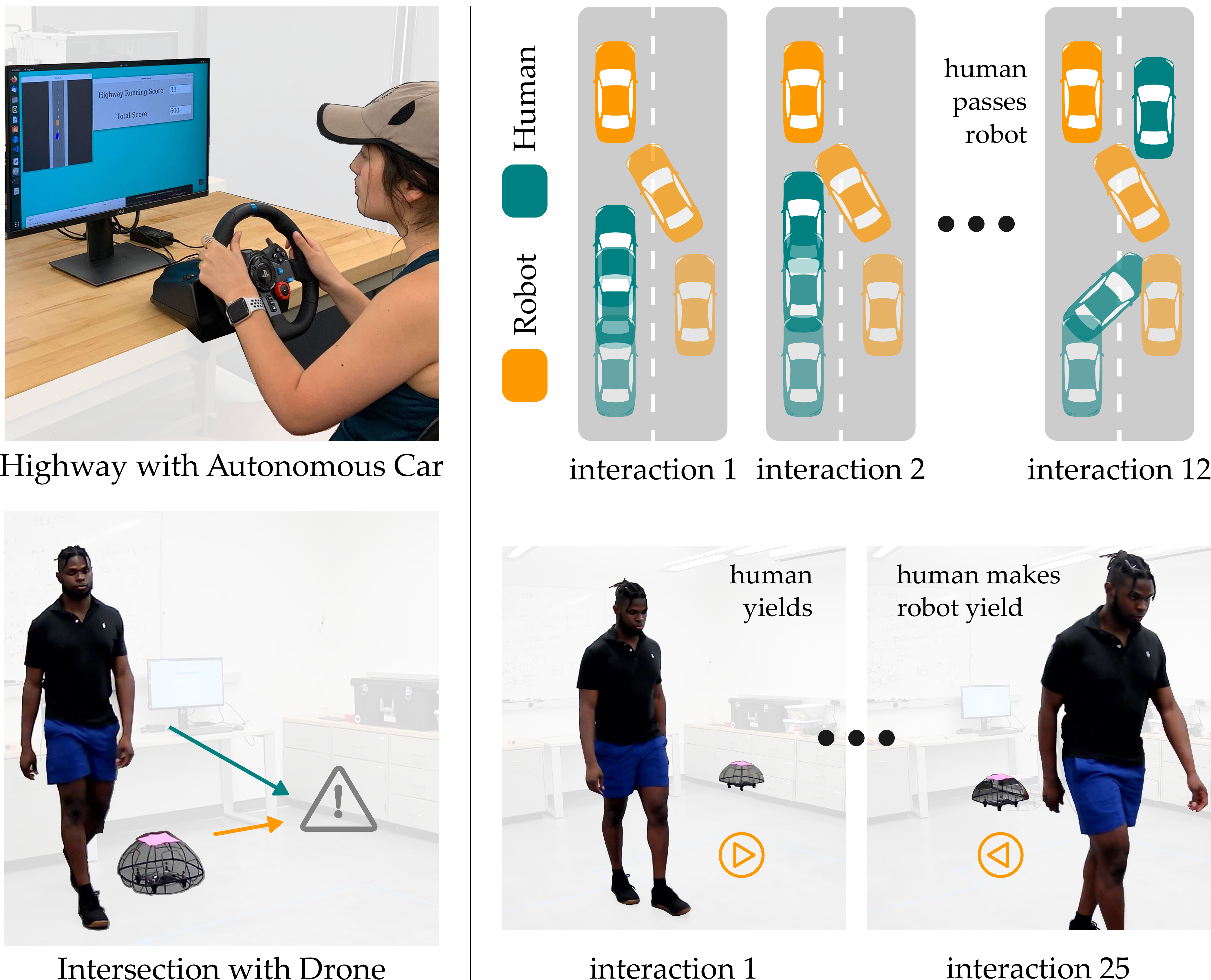}
		\caption{Human interacting with autonomous car (Top) and drone (Bottom). The robot uses state-of-the-art algorithms to influence the human to yield. These approaches work in the \textit{short-term}, but not in the \textit{long-term}.}
		\label{fig:front}
	\end{center}
	\vspace{-2em}
\end{figure}

Consider a human that is driving alongside an autonomous car or walking near a delivery drone (\fig{front}). The human and robot each have their own objectives: perhaps the human wants to drive home as quickly as possible while the autonomous car is trying to ensure that all vehicles share the road safely. During human-robot interaction intelligent robots can leverage their actions to \textit{influence} humans. For instance, here the autonomous car can merge in front of the human to cause this driver to slow down. Merging in front of speeding humans may influence these people the first few times the human and robot interact. But as the human becomes more familiar with the robot's behaviors and capabilities, these actions no longer have the intended effect: over time, human drivers \textit{anticipate} that the autonomous car will change lanes, and \textit{adapt} to avoid the robot or pass it altogether.

Within this paper we define influence as robot actions that (a) emerge as part of the robot's optimal policy and (b) cause nearby humans to change behavior. Today's robots intentionally select influential actions to increase the team's overall reward \cite{sadigh2016planning, tian2022safety, fisac2019hierarchical, ratliff2018perspective, schwarting2019social}, guide humans towards goal regions \cite{bestick2016implicitly, xie2020learning, parekh2022rili, newman2020examining}, or change leader and follower roles \cite{li2021influencing, reily2020leading}. However, these state-of-the-art influencing algorithms often assume that the human is \textit{static}; i.e., the human always interacts with the same robot actions in the same way. For example, some approaches \cite{xie2020learning, parekh2022rili} assume that the will human react to the robot using consistent rules; other methods \cite{sadigh2016planning, tian2022safety, bestick2016implicitly, fisac2019hierarchical, ratliff2018perspective, schwarting2019social} assume that the human treats interaction as a turn-based game, and chooses the optimal response to the robot's behavior.

Prior works indicate that current approaches influence the human as desired in the \textit{short-term}. But we recognize that people are dynamic: over repeated interactions humans will inevitably learn from the robot and adapt their responses. In this paper, we experimentally demonstrate that one common framework for influencing algorithms fails in the \textit{long-term}. We then take a step towards addressing this issue:
\begin{center}\vspace{-0.4em}
\textit{Robots better maintain influence over \emph{long-term} \\ interaction by making their behaviors less \emph{predictable}}.\vspace{-0.4em}
\end{center}
Let us return to our driving example from \fig{front}. When the robot assumes the human is static, it always selects the same influential actions (e.g., merging into the left lane). But as the human gets better at anticipating these behaviors, they become increasingly ineffective (e.g., the human changes lanes and passes on the right). Applying our insight, we envision autonomous systems that --- like their human counterparts --- consistently interact in slightly different ways, making it challenging for humans to predict exactly what the robot will do next. Towards this end, we formalize game-theoretic controllers that optimize for unpredictable and influential robot behaviors across long-term interaction.

Overall, we make the following contributions:

\noindent \textbf{Testing Influencing over Repeated Interaction.} We conduct an online and in-person study where users repeatedly drive alongside an autonomous car. This car solves a Stackelberg game to influence the human: we find that participants are influenced to yield at the start of the experiment, but over time people yield to the robot less frequently.

\p{Formulating Unpredictable Influence} We introduce three possible modifications to an existing game-theoretic framework for influencing humans. These modifications bias the robot's optimal behavior towards actions that purposely obscure the robot's reward function and intended behavior.

\p{Maintaining Influence} We test our approach in driving simulations and an in-person experiment with delivery drones. As participants move across the room their path intersects with the drone: over $25$ interactions, we measure how often the human is influenced to yield. Our results suggest that unpredictable behaviors improve long-term influence.

\section{Related Work}


\noindent\textbf{Influential Actions.}
While robots can also influence humans through social factors such as their expressions or appearance \cite{saunderson2019robots, siegel2009persuasive, rae2013influence}, we here focus on leveraging \textit{actions} to influence humans. Influential actions naturally emerge when robots are interacting with humans and the robot must shape the human's behavior to complete its own task or maximize its own reward \cite{sadigh2016planning, tian2022safety, fisac2019hierarchical, schwarting2019social, parekh2022rili}. Consider a \textit{collaborative} robot arm that is handing cups to a human. How the robot passes these cups will change the human's grasp; as such, the robot orients its cups to guide humans towards more stable grasps \cite{bestick2016implicitly}. Alternatively, take a \textit{competitive} robot arm that is playing air hockey against a human. The actions this robot makes to block the human's shots will alter how the human shoots next time: here robots learn to block in ways that lead opponents towards more easily stopped shots \cite{xie2020learning, parekh2022rili}. Other prior works research influential actions in autonomous driving scenarios that are not necessarily collaborative or competitive \cite{sadigh2016planning, tian2022safety, fisac2019hierarchical, schwarting2019social, hu2022active}. Within this application robots actively guide human drivers towards synergistic behaviors; e.g., an autonomous car nudges into a busy lane so that the humans yield and the autonomous car can seamlessly merge.

Across each of these examples the robot selects influential actions while assuming the human will respond using fixed rules or static patterns. In practice, however, human behavior shifts over time as people learn from and adapt to robots. Unlike prior works, we therefore explore how robots should select influential actions over \textit{long-term} interaction.

\p{Influence \& Game Theory} How do robots identify influential actions in the first place? One common approach is to formulate human-robot interaction as a multi-agent system \cite{schwarting2019social, mehr2021maximum}, and then leverage \textit{game-theoretic} approaches to find robot policies that influence the human towards advantageous behaviors. More specifically, works including \cite{sadigh2016planning, fisac2019hierarchical, tian2022safety} treat human-robot interaction as a Stackelberg game \cite{von2010market} where the robot \textit{acts} (i.e., the robot chooses its actions first) and then the human \textit{reacts} (i.e., the human selects their response given the robot's chosen behavior). Optimal robots in these Stackelberg games intentionally take actions that maximize the robot's reward by shaping the human's response \cite{ratliff2018perspective, foerster2018learning}. But while these game-theoretic approaches generate influential actions, they miss out on: (a) humans are not static agents that always react in the same way and (b) humans and robots act simultaneously, not in turns. As we extend these approaches towards long-term interaction, we seek to capture the human's dynamics and adaption during interaction.
\section{Existing Approaches to Influence}
\label{sec:problem}

\begin{figure*}[t]
    \vspace{0.42em}
	\begin{center}
		\includegraphics[width=1.8\columnwidth]{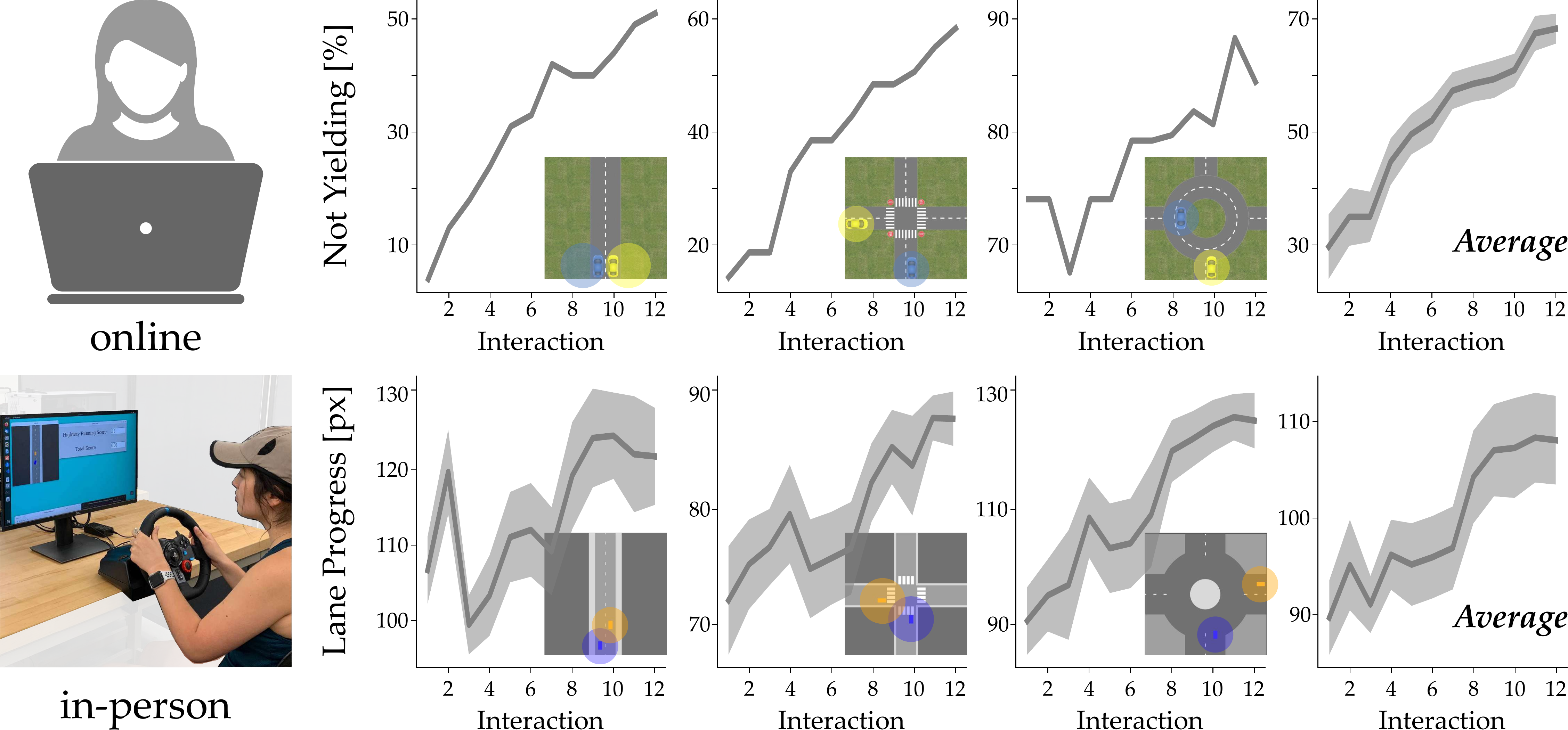}
		\caption{Participants repeatedly interact with an autonomous car that uses existing Stackelberg game approaches to influence their behavior. The autonomous car selects actions $\mathbf{u}_{\mathcal{R}}$ by solving \eq{stackelberg_game}; this is consistent with prior works \cite{sadigh2016planning, fisac2019hierarchical, tian2022safety, ratliff2018perspective, schwarting2019social}. The robot is rewarded for influencing the human to slow down, yield and reduce lane progress. For both online (Top) and in-person (Bottom) participants, \textit{the robot's influence decreases over time}. In the last column (Right) we display the average behavior across the highway, intersection, and roundabout driving environments. Shaded regions show standard error.}
		\label{fig:us1}
	\end{center}
	\vspace{-2.27em}
\end{figure*}

Building on prior works \cite{sadigh2016planning, fisac2019hierarchical, tian2022safety, ratliff2018perspective}, we model human-robot interaction as a discrete-time, general-sum Stackelberg game \cite{von2010market}. This formalism recognizes that the human and robot are trying to complete their own tasks, but each agent's ability to do so is inherently coupled with the other agent's behavior. For example, in the highway scenario from Figure \ref{fig:front} the robot's ability to merge without colliding depends on the human yielding to the robot; in turn, the human sacrifices how fast they can drive home to prevent this collision. Here we will restrict ourselves to interactions between one human and one robot, but this same modelling paradigm extends to an arbitrary number of agents \cite{schwarting2019social}. 

Let $s$ be the system state (e.g., the position of both human and robot cars), let $u_\mathcal{R}$ be the robot's action, and let $u_\mathcal{H}$ be the human's action (e.g., their steering and acceleration). The human-robot system transitions according to the discrete-time dynamics $s^{t+1} = f(s^t, u^t_\mathcal{R}, u^t_\mathcal{H})$. At each timestep $t$ the robot receives reward $r_\mathcal{R}(s^t, u^t_\mathcal{R}, u^t_\mathcal{H})$ and the human receives reward $r_\mathcal{H}(s^t, u^t_\mathcal{R}, u^t_\mathcal{H})$. The human and robot may have different reward functions --- in our running example the human wants to go as quickly as possible while the robot is rewarded for keeping other drivers below the speed limit. 
Let the robot's action trajectory over $T$ timesteps be $\mathbf{u}_{\mathcal{R}} = (u_{\mathcal{R}}^0, \hdots, u_{\mathcal{R}}^{T})$ and let $\mathbf{u}_{\mathcal{H}} = (u_{\mathcal{H}}^0, \hdots, u_{\mathcal{H}}^{T})$ be the human's action trajectory. 
In practice, the human and robot act simultaneously.
But under the Stackelberg game model we separate each interaction into turns: first the robot selects $\mathbf{u}_{\mathcal{R}}$ and then the human responds with $\mathbf{u}_{\mathcal{H}}$. More formally, the robot and human perform bi-level optimization:
\begin{equation}
    \begin{aligned}
        \mathbf{u}_{\mathcal{R}}^* &= \arg \max_{\mathbf{u}_{\mathcal{R}}} ~R_{\mathcal{R}}(s^0, \mathbf{u}_{\mathcal{R}}, \mathbf{u}_{\mathcal{H}}^*(s^0, \mathbf{u}_{\mathcal{R}})). \\
        & \text{s.t.} \quad s^{t+1} = f(s^t, u^t_\mathcal{R}, u^t_\mathcal{H}) \\
        & ~~~ \quad  \mathbf{u}_{\mathcal{H}}^*(s^0, \mathbf{u}_{\mathcal{R}}) = \arg \max_{\mathbf{u}_{\mathcal{H}}} ~R_{\mathcal{H}}(s^0, \mathbf{u}_{\mathcal{R}}, \mathbf{u}_{\mathcal{H}}) \\ 
        & \quad \quad \quad \quad \quad \quad \quad \text{s.t.} \quad s^{t+1} = f(s^t, u^t_\mathcal{R}, u^t_\mathcal{H}) \\
    \end{aligned}
    \label{eq:stackelberg_game}
\end{equation}
Here $R_{\mathcal{R}}(s^0, \mathbf{u}_{\mathcal{R}}, \mathbf{u}_{\mathcal{H}}) = \sum^T_{t=0} r_\mathcal{R}(s^t, u^t_\mathcal{R}, u^t_\mathcal{H})$ is the robot's total reward and $R_{\mathcal{H}}(s^0, \mathbf{u}_{\mathcal{R}}, \mathbf{u}_{\mathcal{H}}) = \sum^T_{t=0} r_\mathcal{H}(s^t, u^t_\mathcal{R}, u^t_\mathcal{H})$ is the human's total reward. When solving \eq{stackelberg_game} both the robot and human maximize their own cumulative reward, but the robot gets to select its actions first.

Robots that apply this formulation use their actions to influence humans. For instance, in \cite{sadigh2016planning} an autonomous car solving \eq{stackelberg_game} nudges into the human's lane to cause the driver to yield, or backs up at an intersection to encourage the human to proceed first. Other state-of-the-art methods have modified the Stackelberg game approach. This includes adding an additional reward term for gathering information about the human's internal state \cite{sadigh2016information}, and parameterizing each agent's reward function with their Social Value Orientation \cite{schwarting2019social}. Other works recognize that humans are not always optimal \cite{fisac2019hierarchical, hu2022active}, and infer whether the human is playing first or second within the Stackelberg game \cite{tian2022safety}.

In this paper we use the Stackelberg game in \eq{stackelberg_game} as our baseline for generating influential robot behaviors. Although previous research has shown that this model influences humans in the \textit{short-term}, we will explore how humans respond over repeated, \textit{long-term} interactions.

\section{Are Humans Influenced in the Long-Term?} \label{sec:user1}

We first performed online and in-person user studies to test whether existing Stackelberg game approaches consistently influence humans during long-term interaction. Participants drove a simulated car while sharing the road with an autonomous vehicle that selected actions according to \eq{stackelberg_game}. Each participant interacted with the autonomous car across three driving settings and $36$ total trials. Our results from $45$ online users and $10$ in-person drivers show that the robot successfully influenced people to yield at first, but over time human drivers adapted to \textit{ignore} or \textit{avoid} the robot.

\p{Experimental Setup} Participants shared the road with an autonomous car in three settings: highway, intersection, and roundabout (see \fig{us1}). 
To simulate the driving environment and vehicle dynamics in real-time we used CARLO \cite{cao2020reinforcement}. 
\textit{In-person} participants controlled their car using a Logitech G29 steering wheel and responsive pedals. 
Each interaction ended after a fixed number of timesteps. \textit{Online} participants first watched an animated video of the start of the interaction, and then selected their behavior from a multiple choice menu. 
Both in-person and online participants earned points for avoiding a collision, staying on the road, and making lane progress. We displayed the participant's current score throughout the experiment. All participants interacted within the highway, intersection, and roundabout settings $12$ times each for a  total of $36$ repeated interactions. 
The road setting order was randomized and balanced across all users.

\p{Independent Variables} The autonomous car solved the Stackelberg game in \eq{stackelberg_game} to select its actions $\mathbf{u}_{\mathcal{R}}$. We rewarded the robot for avoiding collisions and minimizing the human's lane progress. More specifically, we selected:
\begin{equation} \label{eq:U1}
    r_\mathcal{R}(s, u_\mathcal{R}, u_\mathcal{H}) = - \dot{s}_\mathcal{H} - 10 \cdot \mathbb{1}\{\text{collision in }s\}
\end{equation}
where $\dot{s}_\mathcal{H} \subset s$ is the human car's velocity. The robot assumed that the human's reward matched their displayed score:
\begin{multline} \label{eq:U2}
    r_\mathcal{H}(s, u_\mathcal{R}, u_\mathcal{H}) = \dot{s}_\mathcal{H} - 10 \cdot \mathbb{1}\{\text{on road in }s\} - \\ 100 \cdot \mathbb{1}\{\text{collision in }s\}
\end{multline}
Positive values for $\dot{s}_\mathcal{H}$ indicate that the human's car is moving forward along the road (i.e., making lane progress), while negative values mean the human's car is in reverse. Combining Equations (\ref{eq:stackelberg_game})--(\ref{eq:U2}), the robot attempts to \textit{influence} humans to \textit{yield} in order to reduce their lane progress.

\p{Dependent Variables} For \textit{online} participants we recorded whether the human chose to yield or pass the autonomous car. For \textit{in-person} subjects we measured their lane progress, i.e., the vertical distance they traveled. In each environment the human's car started at the bottom of the screen and drove towards the top of the screen; a driver that never yields to the autonomous car would maximize their lane progress.

\p{Participants} For the \textit{online} component of the user study we recruited $63$ anonymous participants. At the start of the experiment these participants read the instructions and then answered qualifying questions to check that they understood the experimental procedure. A total of $45$ users passed these questions and continued on to the survey.

For the \textit{in-person} component we recruited $12$ participants from the Virginia Tech community. Of these, $10$ answered the qualifying questions correctly and completed the experiment ($5$ female, ages $24.7 \pm 5.2$ years). All participants provided informed written consent consistent with university guidelines (IRB \#$20$-$755$). We recognize that users may adapt to become better drivers as they continue to interact in our simulated environment. To account for this confounding factor we had participants practice driving without any autonomous cars until they reached expert-level scores.

\p{Hypothesis} We hypothesized that:
\begin{quote}
\p{H1} \textit{Over repeated interactions the autonomous car's influence will decrease and human drivers will yield to the robot less frequently.}
\end{quote}

\p{Results} Our results from this first user study are summarized in \fig{us1}. The top row shows pass $\%$ for online users as a function of interaction number; the bottom row displays lane progress for in-person drivers over repeated interactions. 

\textit{Online} users chose to either yield or pass the autonomous car during each interaction. We performed Wilcoxon signed-rank tests to see how the human's choice evolved between the first interaction and the final interaction. Our results averaged across all three driving scenarios reveal that humans passed the autonomous car more frequently by the end of experiment ($Z = -5.798$, $p < .001$). This change was also statistically significant for the highway and intersection, 
but not for the roundabout ($Z = -1.155$, $p = .248$). Within the roundabout humans rarely yielded to the robot, perhaps because they perceived their own car as having the right of way.

For \textit{in-person} drivers we measured their lane progress. Remember that the autonomous car is trying to influence humans to reduce their speed; as such, higher lane progress is correlated with less robot influence. Paired t-tests show the human's average lane progress was significantly higher at the final interaction as compared to their first interaction ($t(29) = -5.952$, $p < 0.001$). This trend is consistent across highway ($t(9) = -2.4, p < .05$), intersection ($t(9) = -3.1, p < .05$), and roundabout ($t(9) = -5.6, p < .001$).

Our results from this first study support \textbf{H1}. Autonomous cars that leverage an existing game-theoretic framework to generate influential behaviors are effective in the \textit{short-term}, but do not maintain the same influence across the \textit{long-term}.

\section{Influential and Unpredictable Robots} \label{sec:method}

\begin{figure*}[t]
    \vspace{0.42em}
	\begin{center}
		\includegraphics[width=2.0\columnwidth]{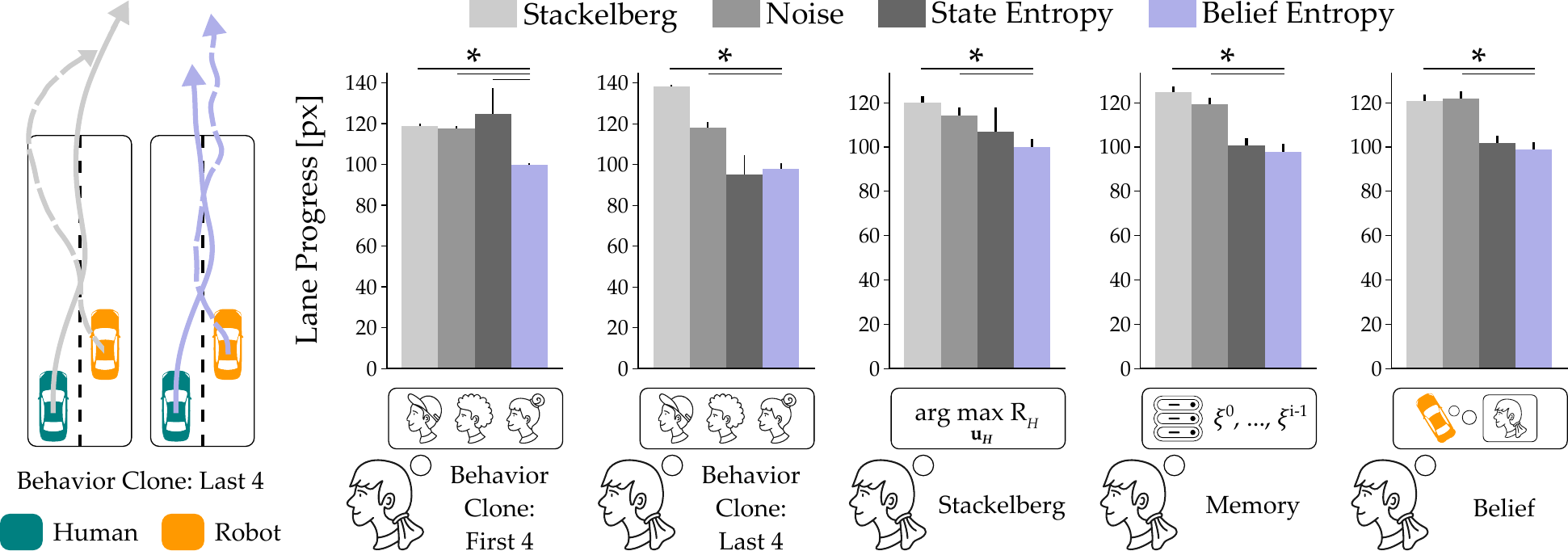}
		\caption{Simulated humans drive alongside an influential robot on the highway. The autonomous car takes actions to reduce the human's lane progress. We test four robot control algorithms: \textbf{Stackelberg} is the state-of-the-art approach from Section~\ref{sec:problem}, while \textbf{Noise}, \textbf{State Entropy}, and \textbf{Belief Entropy} are our proposed modifications from Section~\ref{sec:method}. We then paired these controllers with five different types of simulated humans. \textbf{Behavior Cloned} humans were trained to mimic the actions of in-person users from Section~\ref{sec:user1} (either their \textbf{First 4} interactions or their \textbf{Last 4} interactions). \textbf{Stackelberg} humans follow the Stackelberg game formulation from \eq{stackelberg_game}. \textbf{Memory} humans assume that the robot will follow its average trajectory from the last $3$ interactions, while \textbf{Belief} humans infer whether the robot will coordinate with their behaviors. (Left) Example interactions with a \textbf{Stackelberg} robot (gray) and our proposed \textbf{Belief Entropy} robot (purple). (Right) Average lane progress across $100$ simulated humans. Asterisks $*$ denote statistical significance ($p < .05$).}
		\label{fig:sims}
	\end{center}
	\vspace{-2.27em}
\end{figure*}

Our experiments in Section~\ref{sec:user1} show that the state-of-the-art approach to influential robots falls short over long-term interaction. So what can we do to address this challenge? Here we take a first step towards control strategies that \textit{maintain} influence. Remember our original insight: as humans observe the robot and learn to anticipate its actions, it becomes easy for humans to ignore, avoid, or work around the robot. We therefore propose a game-theoretic approach that combines influential and unpredictable behavior. Specifically, we introduce three possible modifications to the framework from Section~\ref{sec:problem}. These modifications i) inject \textbf{noise}, ii) trade-off between influence and \textbf{state entropy}, or iii) trade-off between influence and \textbf{belief entropy}.

\p{Noise} One na\"ive modification is simply to inject noise into the robot's actions. Here the robot still solves \eq{stackelberg_game} to find its action trajectory $\mathbf{u}^*_{\mathcal{R}} = (u_{\mathcal{R}}^{*,0}, \hdots, u_{\mathcal{R}}^{*,T})$, but at each timestep we add zero-mean Gaussian noise:
\begin{equation} \label{eq:M1}
    u_{\mathcal{R}}^t = u_{\mathcal{R}}^{*,t} + \epsilon^t, \quad \epsilon^t \sim \mathcal{N}(0, \Sigma)
\end{equation}
Covariance matrix $\Sigma$ is a tunable hyperparameter. In practice, safe robots should not take noisy or random actions when those actions could lead to low rewards (i.e., noise $\epsilon$ should not cause a collision). Similar to \cite{fridovich2020confidence}, we therefore constrain $\epsilon$ at each timestep so that $r_\mathcal{R}(s^t, u_\mathcal{R}^t, u_\mathcal{H}^*)$ is probabilistically greater than a designer-chosen threshold $\delta$.

\p{Entropy over States} Our second modification is inspired by human behavior. We recognize that humans never perform the same task in the exact same way; e.g., human drivers naturally vary their timing, acceleration, and steering so that the system state is constantly changing. Here we will similarly encourage robots to visit new states during each interaction by balancing between influence and state entropy. Let $\xi(s^0, \mathbf{u}_{\mathcal{R}}, \mathbf{u}_{\mathcal{H}}) := (s^0, \ldots , s^T)$ be the state trajectory induced by robot and human actions. We augment the robot's reward with the Shannon entropy of the trajectory,  $\mathcal{H}(\xi)$:
\begin{equation} \label{eq:M2}
    R_{\mathcal{R}}(s^0, \mathbf{u}_{\mathcal{R}}, \mathbf{u}_{\mathcal{H}}) = \sum^T_{t=0} r_\mathcal{R}(s^t, u^t_\mathcal{R}, u^t_\mathcal{H}) + \lambda \mathcal{H}(\xi)
\end{equation}
where $\lambda \geq 0$ determines the relative importance of state entropy. Under this approach the robot solves the Stackelberg game in \eq{stackelberg_game} while optimizing for \eq{M2}. Because it is intractable to compute $\mathcal{H}(\xi)$ in real time, we approximate entropy using the particle-based estimate from \cite{liu2021behavior}. Specifically, we use: $\mathcal{H}(\xi) \approx \log \| \xi_{c} - \xi \|$, where $\xi_c$ is the closest trajectory to $\xi$ in our trajectory buffer.

\p{Entropy over Belief} In our final modification the robot purposely makes it harder for humans to predict its objectives. Typically we design robots to reveal their intentions  \cite{dragan2013legibility}. But here we propose the opposite: to maintain influence, robots may \textit{mislead} humans so that users cannot fully anticipate the robot's behaviors. Let the robot's reward $r_{\mathcal{R}} = r_{task} + r_{coord}$ contain two terms: a task reward and a coordination reward. For instance, in autonomous driving $r_{task}$ could be slowing the human and $r_{coord}$ could be avoiding a collision. We will focus on how the robot can increase the human's uncertainty about $r_{coord}$ (e.g., whether the robot actually wants to avoid collisions). Let $b^i = P(r_{coord} \mid \xi^0, \ldots, \xi^{i-1})$ be the human's current belief over $r_{coord}$. Consistent with prior works, we assume that the human updates their belief using \cite{luce2012individual, jeon2020reward}:
\begin{equation} \label{eq:M3}
    b^{i+1} \propto b^i \cdot \exp\big( R_{\mathcal{R}}(s^0, \mathbf{u}_{\mathcal{R}}, \mathbf{u}_{\mathcal{H}}))\big)
\end{equation}
Intuitively, this human thinks the robot will make decisions that are consistent with $r_{coord}$ and approximately optimize the robot's reward function. We encourage the robot to select influential actions \textit{now} that will maximize the human's uncertainty over $b$ at the \textit{next} interaction:
\begin{equation} \label{eq:M4}
    R_{\mathcal{R}}(s^0, \mathbf{u}_{\mathcal{R}}, \mathbf{u}_{\mathcal{H}}) = \sum^T_{t=0} r_\mathcal{R}(s^t, u^t_\mathcal{R}, u^t_\mathcal{H}) + \lambda \mathcal{H}(b^{i+1})
\end{equation}
Within this proposed modification robots apply \eq{M3} to model the human's belief at interaction $i$, and then solve \eq{stackelberg_game} with \eq{M4} as the reward fuction. In practice, this leads to robots that select actions $\mathbf{u}_{\mathcal{R}}$ which make the human uncertain about how the robot will coordinate.
\section{Experiments with Simulated Humans} \label{sec:sims}

\begin{figure*}[t]
    \vspace{0.42em}
	\begin{center}
		\includegraphics[width=2\columnwidth]{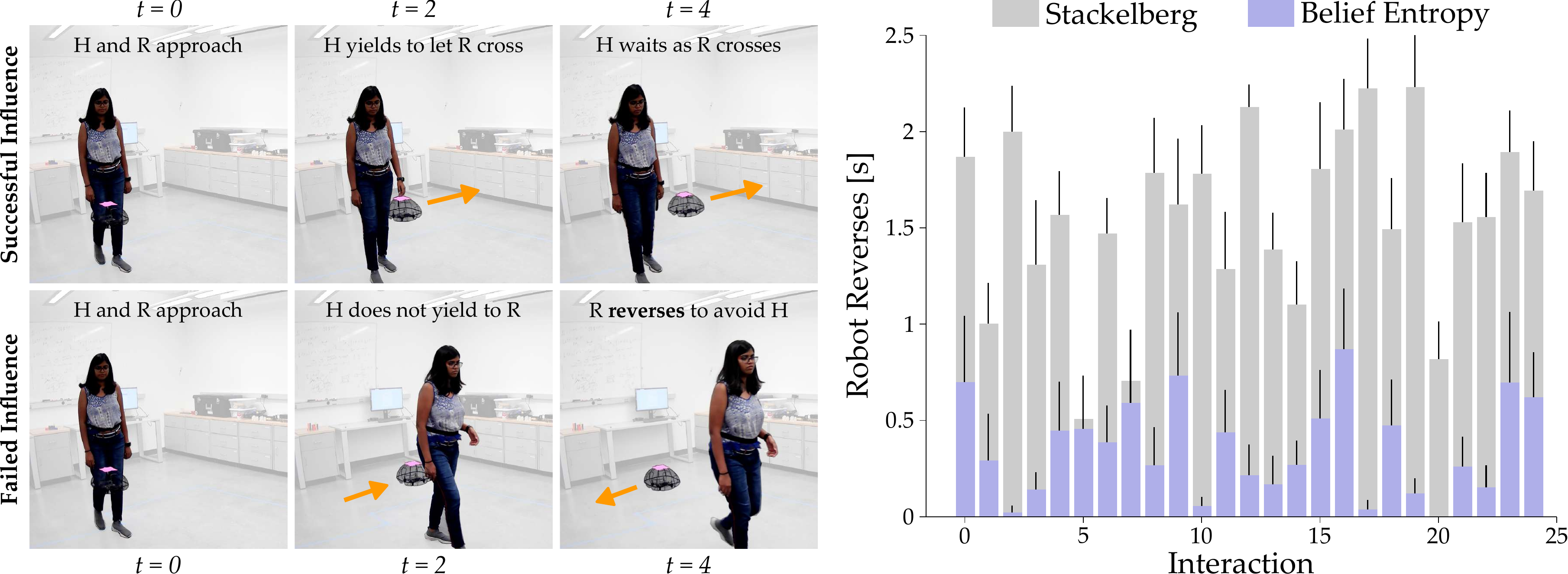}
		\caption{Participants repeatedly cross paths with a drone. (Left) When the drone and human intersect, the drone tries to influence humans to yield so that it can cross first. If this influence failed the robot temporarily reversed direction to avoid a collision. (Right) We plot the amount of time the robot reversed direction across $25$ repeated interactions. The \textbf{Stackelberg} drone solves \eq{stackelberg_game} to try and influence the human; this robot always yielded once the human was within a specific radius. By contrast, the \textbf{Belief Entropy} drone used our modification in \eq{M4} to make its actions less predictable. This drone would occasionally yield at a larger radius or a smaller radius as compared to the \textbf{Stackelberg} drone. We found that this unpredictability reduced the amount of time the robot had to reverse and increased the number of successful influences. Error bars show standard error.}
		\label{fig:us2}
	\end{center}
	\vspace{-2.27em}
\end{figure*}

To test our proposed approaches for long-term influence we first performed experiments with simulated humans (see \fig{sims}). These simulated humans drove alongside an autonomous car on the highway environment from Section~\ref{sec:user1}. The autonomous car's objective matched \eq{U1} in the previous user study: the robot car tried to influence humans to slow down and reduce their lane progress. We combined different types of simulated humans with each influential algorithm. Across these experiments, we found that robots which are \textit{influential and unpredictable} were better able to influence simulated humans and reduce their speed.

\p{Robot Controllers} We implemented four different types of autonomous cars. The \textbf{Stackelberg} baseline selects actions $\mathbf{u}_{\mathcal{R}}$ by solving \eq{stackelberg_game}. \textbf{Noise}, \textbf{State Entropy}, and \textbf{Belief Entropy} are our proposed modifications. None of the autonomous cars knew what type of simulated human they would be interacting with.

\p{Simulated Humans} Actual human drivers exhibit a variety of different behaviors. To try and mimic this diversity, we designed five different simulated humans. We trained two \textbf{Behavior Cloning} models using the highway data from our in-person study in Section~\ref{sec:user1}. The first model was trained on data from the \textbf{first 4} interactions (where users were more influenced by the robot), and the second model was trained on data from the \textbf{last 4} interactions (where users yielded to the robot less frequently). Next, we simulated a \textbf{Stackelberg} human that behaves according to \eq{stackelberg_game}. In practice, humans may assume that the robot will behave the same way this interaction as it did on the previous interaction. We therefore designed \textbf{Memory}, a simulated human that predicts the robot will match its average trajectory from the last $3$ interactions. Finally, the \textbf{Belief} human uses \eq{M3} to infer whether the robot will coordinate with the human (e.g., if the robot will move out of the way to avoid a collision).

\p{Dependent Measures} We paired each combination of human and robot and then simulated every pair for $100$ interactions. The initial state of the cars was randomized, but we ensured that the autonomous car always started ahead of the human's car. At the end of each interaction we measured the human car's total lane progress (in pixels).

\p{Results} Our results from this simulation are displayed in \fig{sims}. On the left we show example trajectories the cars followed, and on the right we plot average lane progress. To analyze these results we performed repeated measures ANOVAs. We found that the robot's controller had a significant main effect on lane progress: the results of post hoc analysis are highlighted in \fig{sims}. For each type of simulated human the \textbf{Belief Entropy} approach resulted in significantly lower lane progress (as compared to \textbf{Stackelberg} and \textbf{Noise}). These results support our proposed modifications: robots that purposely make their behaviors more unpredictable are better able to influence our array of simulated humans.

\section{User Study} \label{sec:user2}

Our simulations supported our modifications (and in particular \textbf{Belief Entropy}). We therefore conducted a second study in which real humans interacted with a drone (see \fig{front}). Here the human and robot repeatedly intersected each other's path: at these intersections, the drone flew forward to attempt to influence humans to yield. Over $25$ repeated interactions, we measured whether the drone was able to maintain influence and keep the right-of-way. We compared a proposed framework for influential but unpredictable robots to the state-of-the-art Stackelberg game formalism.

\p{Experimental Setup} Participants shared space with a drone (see \fig{us2}). We tracked the drone using ceiling-mounted cameras, and humans wore an HTC Vive Tracker around their waist for real-time position measurements. Participants walked back and forth across the room to pick up blocks and build a tower; each time the human started to cross, the drone moved orthogonally to intersect with the human's path.

\p{Independent Variables} We compared two robot controllers: the \textbf{Stackelberg} baseline from Section~\ref{sec:problem} and our proposed \textbf{Belief Entropy} modification from \eq{M4}. The robot was rewarded for crossing the room as quickly as possible while avoiding collisions with the human: to maximize its speed, the robot tried to influence humans to yield. The drone selected actions in real-time using the reward functions:
\begin{equation} \label{eq:U2_1}
    r_\mathcal{R}(s, u_\mathcal{R}, u_\mathcal{H}) = \dot{s}_{\mathcal{R}} - 10 \cdot \mathbb{1}\{\text{collision in }s\}
\end{equation}
\begin{equation} \label{eq:U2_2}
    r_\mathcal{H}(s, u_\mathcal{R}, u_\mathcal{H}) = \dot{s}_{\mathcal{H}} - 100 \cdot \mathbb{1}\{\text{collision in }s\}
\end{equation}
where $\dot{s}_{\mathcal{R}} \subset s$ is the robot's forward velocity and $\dot{s}_{\mathcal{H}} \subset s$ is the human's velocity. Negative values for $\dot{s}_{\mathcal{R}}$ indicate that the drone is reversing direction and yielding to the human. Recall that the \textbf{Belief Entropy} robot takes actions that make the human uncertain about whether the robot will coordinate (i.e., whether the robot is optimizing for avoiding collisions). In practice, this caused the \textbf{Belief Entropy} drone to randomly switch between crossing aggressively (only yielding if the human was within a small radius) and defensively (yielding if the human was anywhere within a larger radius).

\p {Dependent Measures} A robot that maintains the right-of-way will always have a positive $\dot{s}_{\mathcal{R}}$. However, if the human insists on going first, then the robot must back off and give the participant space. To measure influence, we therefore recorded the amount of time the robot \textit{reversed} during each interaction. Lower values correspond to higher influence.

\p {Participants} We recruited $11$ participants from the Virginia Tech community ($10$ male, ages $22.1 \pm 3.1$ years). These participants provided informed consent under IRB \#$20$-$755$. We recognized that people may hesitate to walk close to a flying drone; we accordingly demonstrated the task and drone behaviors before starting the experiment. We leveraged a within-subjects design: all participants interacted with a \textbf{Belief Entropy} robot $25$ times and a \textbf{Stackelberg} robot $25$ times. The order of presentation was balanced across users.

\p{Hypothesis} We hypothesized that:
\begin{quote}
\p{H2} \textit{A drone that optimizes for influential but unpredictable actions will better maintain influence than a purely influential drone.}
\end{quote}

\p {Results} Our results are summarized in \fig{us2}. Paired t-tests reveal that the \textbf{Stackelberg} robot spent significantly more time backing-up and yielding to the human as compared to \textbf{Belief Entropy} ($t(327) = 13.02, p < .001$). We also noticed that --- as participants became more familiar with the \textbf{Stackelberg} robot --- they insisted on going first more frequently (perhaps because they were able to predict when this robot would yield). As a result, the final $8$ interactions with \textbf{Stackelberg} had a higher average reverse time than the first $8$ interactions, although this difference was not statistically significant ($t(10) = -1.45, p = .09$). Overall, the findings from our second study supported \textbf{H2}: the \textbf{Belief Entropy} robot consistently influenced the $11$ participants to yield the right-of-way over $25$ repeated interactions.
\section{Conclusion}

Our work is a step towards long-term human-robot interaction. We first demonstrated that a prevalent framework for influential robots is effective in the short-term, but humans adapt to these influential actions over repeated interactions. We next proposed three modifications to make the robot's behaviors less predictable. Our simulations and experiments support these modifications, and indicate that robots which are less predictable may be more influential in the long-term.


\newpage
\balance
\bibliographystyle{IEEEtran}
\bibliography{IEEEabrv,bibtex}

\end{document}